# Modeling and analysis of pHRI with Differential Game Theory

Paolo Franceschi[1], Manuel Beschi[2,3], Nicola Pedrocchi[3], Anna Valente[1]

*Abstract*— Applications involving humans and robots working together are spreading nowadays. Alongside, modeling and control techniques that allow physical Human-Robot Interaction (pHRI) are widely investigated. To better understand its potential application in pHRI, this work investigates the Cooperative Differential Game Theory modeling of pHRI in a cooperative reaching task, specifically for reference tracking. The proposed controller based on Collaborative Game Theory is deeply analyzed and compared in simulations with two other techniques, Linear Quadratic Regulator (LQR) and Non-Cooperative Game-Theoretic Controller. The set of simulations shows how different tuning of control parameters affects the system response and control efforts of both the players for the three controllers, suggesting the use of Cooperative GT in the case the robot should assist the human, while Non-Cooperative GT represents a better choice in the case the robot should lead the action. Finally, preliminary tests with a trained human are performed to extract useful information on the real applicability and limitations of the proposed method.

## I. INTRODUCTION

In today's industries, there is a significant interest in deploying collaborative applications involving a human operator and a robot that work together towards a shared objective. In particular, *collaboration* happens when the human operator and the robot must execute a task together, and the action of the one has immediate consequences on the other. More precisely, it is defined as physical Human-Robot Interaction (pHRI) [1] when communication occurs through interaction forces.

A widespread technique to manage the pHRI is Impedance Control (IC), as discussed in [2]. Adaptive IC for pHRI can be classified as an adaptation of the impedance set-point ([3]) and an adaptation of the mass-spring-damper parameters ([4]). Moreover, hybrid controllers that combine both can be defined as in [5], [6]. Among the models proposed to describe pHRI, Game Theory (GT) provides valuable tools to analyze complex interactive behaviors involving multiple agents, as discussed in [7]. GT provides models (cooperative, non-cooperative, multi-stages, etc.) of strategic interaction among decision-makers. It provides the players with "optimal" policies to minimize their objectives, taking into account interaction.

Previous studies investigated the interaction between humans and machines in a non-cooperative game-theoretical (NCGT) framework. [8], and [9] studied an individual's behavior and two individuals approaching classical GT problems. It shows that individual players tended towards a cooperative solution, whereas two players tended towards a Nash equilibrium on average. The cooperative solution would be the best, except when they do not trust each other. [10] presents results where three out of six individuals behave more in a non-cooperative mood rather than optimal control in interaction with a programmed active front steering (AFS). In this work, the AFS is programmed to behave as a rational player in a non-cooperative framework. [11] presents interesting results of a human dyad that performs shared reference tracking and shows that their behavior is better described by the non-cooperative model rather than a model that considers the partner's action as a system disturbance. All these works investigate the non-cooperative case only, putting the participants in this particular situation (human dyads are not allowed to communicate, and the machines the humans are interacting with are programmed in a non-cooperative way). There exist cases that investigate the modeling of a cooperative scenario [8], [12], and in [13], the cooperative model is among the four presented models of interaction between a human and a steering mechanism.

GT has also found applications for the pHRI. [14] uses the Nash Equilibrium to update the robot cost function according to the interaction force, resulting in a variable impedance control with damping and stiffness updated online. The work in [15] solves the same differential non-cooperative game-theoretic problem differently by a policy iteration. Finally, a framework for differential game-theoretic modeling of HRI is in [16] proposing an observer for each other's control laws to address different behaviors. Such modeling is extended in [17] for trajectory tracking in the non-cooperative scenario.

On the one hand, such works demonstrate that differential (continuous-time) non-cooperative models provide optimal behavior for non-cooperative HRI. On the other hand, when the interaction is physical, there is a dramatic margin of improvement in exploiting cooperative Game-Theory. The solutions of the non-cooperative games given by the Nash Equilibrium are frequently not optimal, and the agents' cooperation improves payoffs to all players [18].

This work investigates the Cooperative solution, and different conditions are considered. Namely, the human knows the robot is cooperating, while the robot is programmed according to the cooperative solutions. Indeed, cooperative

[1]Paolo Franceschi and Anna Valente are with the Department of Innovative Technologies, University of Applied Science and Arts of Southern Switzerland (SUPSI), Lugano, Switzerland {*name.surname*}`@supsi.ch`

[2]Manuel Beschi is with the Università di Brescia, Dipartimento di Ingegneria Meccanica ed Industriale, Brescia, Italy `manuel.beschi@unibs.it`

[3]Nicola Pedrocchi and Manuel Beschi are with the Institute of Intelligent Industrial Technologies and Systems for Advanced Manufacturing of the National Research Council of Italy (CNR-STIIMA), Milano, Italy. {*name.surname*}`@cnr.stiima.it`

solutions are typically better than non-cooperative ones if both players agree to cooperate.

In Cooperative Game Theory (CGT), a crucial theoretical result is in [19], which shows how the Cooperative Differential Game (CDGT) formulation may turn into a quadratic game, solvable through standard control approaches. This work proposes adopting the Linear Quadratic CGT to control a pHRI application in such a context. In [20], it was investigated a CGT formulation for role arbitration between a human and a robot. Role arbitration between humans and robots is an increasing research topic to allow deliberative robotics [21], and the robot's behavior should be described and modeled according to different interacting modes.

This work extends such a model by allowing different reference tracking for the human and the robot, which was not considered in the previous work. Specifically, the paper investigates the tracking of different references of the two players, allowing for arbitration between humans and robots in guidance applications. The equivalence of such a formulation with an IC is shown.

To the best of the authors' knowledge, the solution to the LQ-CGT infinite horizon problem involving the definition of a shared reference is not available in the literature. Previous works are limited to the discrete-time finite-horizon cases [12], [22] or to track the same reference [19]. However, no argumentation on the agreement of a single reference starting from two different references is provided.

While the definition of two references and the agreement on a shared reference might seem useless for practical uses, it represents a typical problem in assistive driving and shared control as [12], [13]. As a clarifying example, consider the case in which a human and a robot should co-transport a significant component. The robot's trajectory is precomputed while moving from the starting point to the target one. The human can reshape it for many reasons (*e.g.,* there is a dynamic obstacle, or simply the human prefers to make it slower). In this case, a shared reference should be considered the robot's target pose. This shared reference must also be weighted based on information the robot knows but the human might not know (*e.g.,* proximity to singularities or joint limits). An example application can be in [23].

This work wants to investigate all these situations where the human is leading, and the robot must assist, or conversely, the robot drives the system far from unwanted situations and all the possible situations in between to have a smooth transition. Specifically, in such an application field, the contributions of this work are: (i) to study the solutions of the LQ-CGT and the LQ-NCGT games to see the game's different behaviors according to different tuning parameters; (ii) to give insights into the game-theoretic description of pHRI tasks and prepares the way to develop adaptive controllers based on the GT formulation, capable of switching from one model to the other according to role arbitration logic; (iii) to analyze such games according to parameter tuning; and (iv) to provide the solution to the infinite horizon LQ-CGT case (as a minor contribution). The proposed method is compared with (i) the Linear Quadratic Regulator (LQR) and (ii) the Non-Cooperative Game-Theoretic (NCGT) controller. The LQR method is selected to show that results differ despite the very similar formulation (see Section II). The NCGT is used to compare the two solutions to GT problems because previous works showed that NCGT represents a suitable interactive formulation for the problem considered in this work. This comparison allows us to understand and analyze the pros and cons, the behavior of the various controllers, and their applicability in real-world scenarios. A set of simulations shows how different tuning of control parameters affects the system response and control efforts of both the players for the three controllers. Finally, experiments show the applicability of the method to pHRI.

## II. METHOD

We model the desired robot motion using a Cartesian Impedance system at the end-effector, as Cartesian space is more natural for the human operator:

$$M_i a(t) + D_i v(t) + K_i \Delta x = u_h(t) + u_r(t) \quad (1)$$

where $M_i$, $D_i$ and $K_i \in \mathbb{R}^{6\times 6}$ are the desired inertia, damping, and stiffness matrices. $\Delta x = x - x_0$ where $x_0$ is the equilibrium position of the virtual spring, and $u_h \in \mathbb{R}^6$ and $u_r \in \mathbb{R}^6$ represent the human and robot effort applied to the system. $x$ is defined as in [24], with the vector $x = [p^T \ \phi^T]^T$ where $p^T$ are the position coordinates and $\phi^T$ the set of Euler angles that defines the rotation matrix describing the end-effector orientation. Also, we can write the vector containing the linear and angular velocities as $v = [\dot{p}^T \ \omega^T]^T$ with $\dot{p}$ the linear velocity, $\omega$ angular velocity. Finally, (1) can be linearized in the state-space formulation around the working point[1]:

$$\dot{z} = Az + B_h u_h + B_r u_r \quad (2)$$

where $z = [\Delta x^T \ v^T]^T \in \mathbb{R}^{12}$ is the state space vector, $A = \begin{bmatrix} 0^{6\times 6} & J_\omega \\ -M_i^{-1}K_i & -M_i^{-1}D_i \end{bmatrix}$, $B_h^{12\times 6} = B_r^{12\times 6} = \begin{bmatrix} 0^{6\times 6} \\ M_i^{-1} \end{bmatrix}$, with $0^{6\times 6}$ denoting a $6 \times 6$ zero matrix.

Typically robot controllers accept reference positions or velocities in the joint space as a control input. Reference velocity in the joint space results in:

$$\dot{q}_{ref}(t) = J(q)^+ v(t), \quad \text{with} \quad \dot{q}_{ref}(t) \in \mathbb{R}^n \quad (3)$$

where $n$ is the number of joints and $J(q)^+$ is the pseudo-inverse of the geometric Jacobian. Simple integration allows commanding joint positions instead of velocities to the robot. Considering that today's robots have excellent tracking performance in the frequency range excitable by the operator, in this work, hypothesis $\dot{q} \simeq \dot{q}_{ref}$ is assumed to hold.

### A. Differential Cooperative Game Theoretic modeling

Rewriting (1) as (2), allows including it in a Differential Cooperative Game Theory (DCGT) framework, as in [25]. For this purpose, the system in (2) can be rewritten as

$$\dot{z} = Az + Bu \quad (4)$$

---

[1]the non-linearity is due to rotations, this work considers small rotations, as pHRI applications (*e.g.,* transportation of large/heavy objects) typically involve large translations and small rotations

with $B \in \mathbb{R}^{12 \times 12} = [B_h \; B_r]$ and $u = [u_h \; u_r]^T \in \mathbb{R}^{12 \times 1}$.

The human and robot targets may differ and are defined as $z_{ref,h}$ and $z_{ref,r}$, respectively. The two agents' objectives can be modeled as the minimization of a cost function,

$$J_h = \int_0^\infty \left( (z - z_{ref,h})^T Q_{h,h} (z - z_{ref,h}) \right. \\ \left. + (z - z_{ref,r})^T Q_{h,r} (z - z_{ref,r}) + u_h^T R_h u_h \right) dt \quad (5)$$

$$J_r = \int_0^\infty \left( (z - z_{ref,h})^T Q_{r,h} (z - z_{ref,h}) \right. \\ \left. + (z - z_{ref,r})^T Q_{r,r} (z - z_{ref,r}) + u_r^T R_r u_r \right) dt \quad (6)$$

where $J_h$ and $J_r$ are the cost that the human and the robot incur, $Q_{h,h}, Q_{h,r} \in \mathbb{R}^{6 \times 6}$ and $Q_{r,h}, Q_{r,r} \in \mathbb{R}^{6 \times 6}$ matrices that weights the state and references and $R_h \in \mathbb{R}^{6 \times 6}$ and $R_r \in \mathbb{R}^{6 \times 6}$ weights on the control input.

Consider the physical meaning of (5) and (6). In pHRI, humans and robots want to minimize their effort, subject to the system dynamics and the opponent's control action. GT admits two different criteria for the computation of the best $u_r$. In NCGT, the Nash Equilibria gives the different local minima solutions. The set of all the solutions is discrete and finite, and typically, NE solutions are not Pareto optimal. Indeed, the function costs of the player could be higher if the players agree to cooperate. This agreement among players happens in the CGT, granting the Pareto optimality for the solutions, *i.e.*, no change leads to cost improvement for some agents without others losing. In CDGT, the solutions lie on a hyper-surface called a Pareto frontier.

In our case, the system evolves linearly with the state and control input, and the cost of each player is a quadratic function of the state and the control input. As described in [25], the multi-goal function costs of the CDGT problem can be turned into a Differential Linear Quadratic problem by defining a weighted cost of the different player's function costs. Modifying the weights allows the selection of an optimal solution on the Pareto frontier. Hence, the cost that the two players minimize together can be written as follows:

$$J_{gt} = \alpha J_h + (1 - \alpha) J_r = \int_0^\infty \left( \tilde{z}^T Q_{gt} \tilde{z} + u^T R_{gt} u \right) dt \quad (7)$$

with $u = [u_h^T, u_r^T]^T$ and $\tilde{z} = z - z_{ref}$. The values for $z_{ref}$, $Q_{gt}$ and $R_{gt}$ must be defined, and $\alpha \in (0, 1)$ represents the weight each player's cost has in the overall cost.

The Linear Quadratic CGT finds the input $u$ as linear feedback on the state that minimizes (7) subject to dynamical constraints. Given (4), the control action $u$ results in:

$$u = -K_{gt} \tilde{z} = -K_{gt} z + K_{gt} z_{ref} \quad (8)$$

with $K_{gt} = R_{gt}^{-1} B^T P$, and $P$ is the solutions of the Algebraic Riccati Equation (ARE)

$$0 = A^T P + P A^T - P B R_{gt}^{-1} B^T P + Q_{gt}$$

The LQ-CGT turns into a classical LQR problem, *i.e.*:

$$\min_u \quad J_{gt} = \int_0^\infty \left( \tilde{z}^T Q_{gt} \tilde{z} + u^T R_{gt} u \right) dt \\ s.t. \quad \dot{\tilde{z}} = A\tilde{z} + Bu \quad \vee \quad \tilde{z}(t_0) = \tilde{z}_0 \quad (9)$$

and combining (5), (6) into (7) (see Appendix):

$$Q_{gt} = \alpha(Q_{h,h} + Q_{h,r}) + (1 - \alpha)(Q_{r,h} + Q_{r,r}) \\ R_{gt} = \begin{bmatrix} \alpha \hat{R}_h & 0^{6 \times 6} \\ 0^{6 \times 6} & (1 - \alpha) R_r \end{bmatrix}$$

Defining $Q_h = \alpha Q_{h,h} + (1 - \alpha) Q_{h,r}$ and $Q_r = \alpha Q_{r,h} + (1 - \alpha) Q_{r,r}$ (see Appendix), the reference $z_{ref}$ results in a weighted composition of the human and robot references,

$$z_{ref} = Q_{gt}^{-1} (z_{ref,h} Q_h + z_{ref,r} Q_r). \quad (10)$$

Given the last three equations, the problem introduced by (9) can then be solved, with $K_{gt}$ representing the feedback gain matrix of the human and the robot, and (10) representing the agreed shared reference for both the human and the robot.

### B. The game as Impedance Control tuning

Consider (8). The feedback gain matrix $K_{gt}$ is composed by two sub-matrices: $K_{gt} = [K_h^T, K_r^T]^T$. Moreover, recalling that the state is composed of positions and velocities, the two matrices have components $K_h = [K_{h,p}^T, K_{h,v}^T]^T$ and $K_r = [K_{r,p}^T, K_{r,v}^T]^T$, with subscripts $p$ and $v$ identifying the components multiplying positions and velocities, respectively. Hence, $u_h$ and $u_r$ can be written as

$$u_h = -K_h(z - z_{ref}) = \begin{bmatrix} -K_{h,p} & -K_{h,v} \end{bmatrix} \begin{bmatrix} \Delta x - x_{ref} \\ v \end{bmatrix} \quad (11)$$

$$u_r = -K_r(z - z_{ref}) = \begin{bmatrix} -K_{r,p} & -K_{r,v} \end{bmatrix} \begin{bmatrix} \Delta x - x_{ref} \\ v \end{bmatrix} \quad (12)$$

Substituting (12) into (1), it results in:

$$M_i a + (D_i + K_{r,v}) v + (K_i + K_{r,p}) \Delta x = u_h + K_{r,p} x_{ref} \quad (13)$$

which is an Impedance Controller equation subject to an external force. Note that, in (13), the damping and stiffness matrices vary according to the robot feedback gain matrix that is a function of the weight parameter $\alpha$.

## III. EXPERIMENTS

The proposed approach is compared in simulations and preliminary experiments with a trained human, with the classical feedback LQR and the feedback NCGT.

*1) LQR case:* the human and the robot aim at minimizing their cost functions without any regard for the other player's actions, which are given by

$$J_{h,lqr} = \int_0^\infty \left( (z - z_{ref,h})^T Q_h (z - z_{ref,h}) + u_h^T R_h u_h \right) dt \quad (14)$$

$$J_{r,lqr} = \int_0^\infty \left( (z - z_{ref,r})^T Q_r (z - z_{ref,r}) + u_r^T R_r u_r \right) dt \quad (15)$$

which have solutions $K_{h,lqr} = R_h^{-1} B_h^T P_{h,lqr}$ and $K_{r,lqr} = R_r^{-1} B_r^T P_{r,lqr}$, where $P_{h,lqr}$ and $P_{r,lqr}$ are solutions of the AREs.

*2) NCGT case:* the human and the robot aim to minimize their cost functions, subject to the other influence:

$$J_{h,nc} = \int_0^\infty ( (z - z_{ref,h})^T Q_h (z - z_{ref,h}) \\ + u_h^T R_{h,h} u_h + u_r^T R_{h,r} u_r ) \, dt \quad (16)$$

$$J_{r,nc} = \int_0^\infty ( (z - z_{ref,r})^T Q_r (z - z_{ref,r}) \\ + u_r^T R_{r,r} u_r + u_h^T R_{r,h} u_h ) \, dt \quad (17)$$

which have solutions $K_{h,nc} = R_h^{-1} B_h^T P_{h,nc}$ and $K_{r,nc} = R_r^{-1} B_r^T P_{r,nc}$, where $P_{h,nc}$ and $P_{r,nc}$ are solutions of coupled AREs. For full treatment, please refer to [25].

For all the cases, the matrices $Q_h$ and $Q_r$ are defined as in the cooperative case.

### A. Simulations

We designed the virtual experiments to show the system response according to the tuning of i) $\alpha$, ii) $Q_{r,r}$; iii) $R_r$. Indeed, the robot cost function can be varied at will, while the human cost function cannot be imposed. The simulation parameters are $z_{ref,h} = 1\,m$; $z_{ref,r} = 0.5m$; state weight matrices $Q_{h,h} = diag([1, 0.0001])$, $Q_{h,r} = diag([0,0])$, $Q_{r,h} = diag([0,0])$ and $Q_{r,r} = diag([1, 0.0001])$; control weight matrices $Rh = 0.0005$ and $Rr = 0.0005^2$. The system (impedance) parameters are $M_i = 10N$, $D_i = 25Ns/m$ and $K_i = 0N/m$. Setting the stiffness coefficient null is common when pHRI is involved, as in [16], [4]. However, this particular stiffness choice is not mandatory, and the method is still valid otherwise. Finally, the control input of the robot represents an additional stiffness, with a stiffness matrix equal to the control gain matrix, as described in II-B.

*1) Simulation 1 ($\alpha$ modification):* Figure 1 shows the position trajectory and the control inputs for three different values of $\alpha = \{0.2, 0.5, 0.9\}$. The LQR (dashed lines) is less affected by the change of $\alpha$ since the other's control input can be seen as an external disturbance, as the player does not know the opponent's influence in the system. NCGT (dotted line), instead, shows a significant variation. In NCGT and LQR, both players keep putting effort even when an equilibrium is reached. In NCGT, no player has an advantage in reducing his effort when Nash equilibrium is achieved. In LQR, an equilibrium is reached against external disturbance, and the player's effort can be seen as a disturbance rejection. Finally, in CGT, both players reduce their effort to zero after reaching the equilibrium point because they reach the shared reference together, and no one is interested in changing it.

*2) Simulation 2 (small $Q_{r,r}$):* Figure 2 shows the case in which $Q_{r,r} = diag([0.1, 0.0001])$, and $\alpha = \{0.01, 0.05, 0.1, 0.2, 0.5, 0.9\}$. The figure plots only CGT variations since the other controllers have similar behaviors to the one presented for the CGT. Figure 2(a) shows the variation in the state according to $\alpha$ by lowering the robot's state weight matrix. Because the weight on the robot's reference is low, the state goes quickly towards the direction of the human's reference even for small values of $\alpha$ ($> 0.1$). The robot reference becomes relevant only with minimal values of $\alpha$. In this simulation, $\alpha = 0.1$ allows for making them comparable, being the robot's state weight one-tenth of the human's, Figures 2(b-c) show the control efforts are similar to Simulation 1 (*e.g.*, see the case $\alpha = 0.5$).

---

[2]These values are computed via Inverse Optima Control techniques on a trained human. The robot is set to mimic the human cost function.

*3) Simulation 3 (small $R_r = \{5e^{-5}, 1e^{-4}, 1e^{-3}\}$):* Figure 3 shows the results by keeping $\alpha = 0.5$. On the one hand, CGT keeps constant the equilibrium position, resulting from an agreement depending on $\alpha$. On the other hand, the human and robot control inputs vary according to $R_r$. The effort of the robot increases as the $Rr$ decreases, and vice versa for the human effort. Conversely, LQR and NCGT reach different equilibrium points according to $R_r$. Indeed, LQR can partially compensate for the human control effort like an external disturbance. Figure 4 shows the costs according to $\alpha$. The human cost of the cooperative case is lower for $\alpha > 0.6$. Increasing $\alpha$ is preferable for the human when he wants to lead because the robot assists more humans in reaching their target. Conversely, low $\alpha$ means that the human is more willing to help the robot reach its target.

*4) Simulations Notes:* CGT minimizes both players' control effort, at least at the equilibrium. CGT is more favorable for humans with high values of $\alpha$, possibly with the human leader. At the same time, NCGT or LQR are more likely to be adopted with low values of $\alpha$ for the robot-leading situation. Low values of $R_r$ allow the robot to be more helpful with high $\alpha$ (human leading) and be more capable of driving the system to its reference with low values of $\alpha$ (robot leading).

### B. Real-world experiments

Experiments are performed with a human operator cooperating with a UR5 robot. Two relevant components of the proposed controller must be addressed when dealing with real humans: knowledge of the human's reference and cost function. An Inverse Optimal Controller (IOC) [26], [27] recovers an estimate of the human parameters, $\hat{Q}_h$ and $\hat{R}_h$. The robot knows the human target that defines the reference in (10). This might seem too restrictive, but it allows fair testing of the controller's performance. Human target pose identification is a well-investigated topic [28], [29].

The humans performing the experiments know the robot's behavior and the different control techniques. This awareness is achieved after proper training, even if the human does not know anything initially. This choice is made for different reasons. (1) the methodology copes with a human aware of the robot, so it is expected that the operator knows (at least approximately) the robot's behavior and has some knowledge. (2) The knowledge of the opponent's intention is among the hypothesis of the GT formulation. The purpose of the experiments is to verify if the CGT describes pHRI and to avoid additional noise. This hypothesis must be relaxed in real-world applications, which will be addressed in future works. (3) If the human is aware of the CGT behavior of the robot, better results can be achieved thanks to cooperation. This experiment can be seen as a test for the method's long-term applicability. This assumption is also in [12].

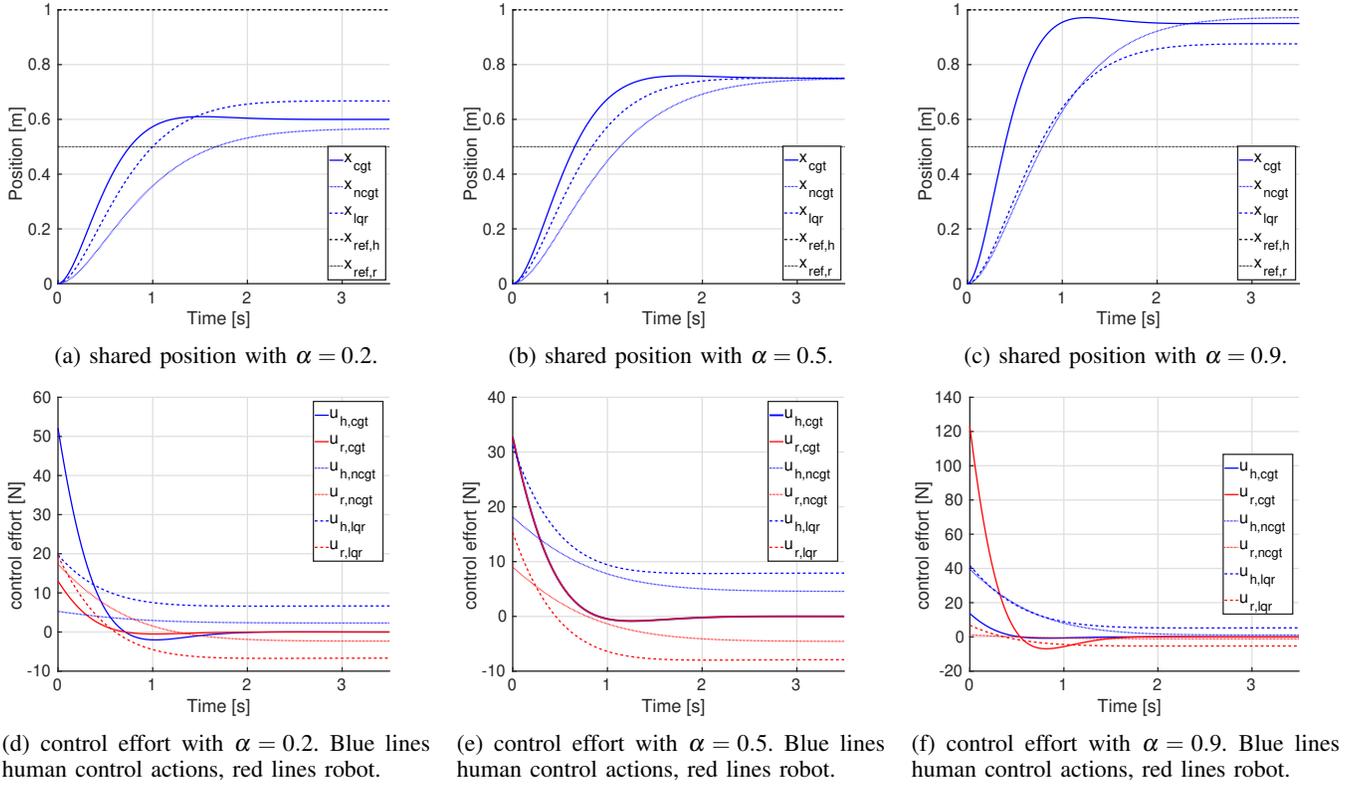

Fig. 1: position and control effort comparison varying weight parameter $\alpha$.

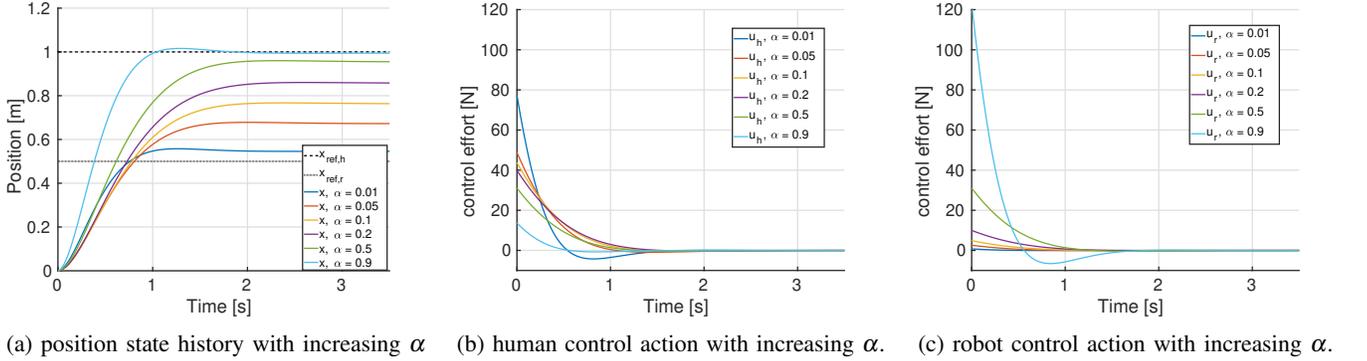

Fig. 2: Position and control effort comparison varying weight parameter $\alpha$ with low weight on the robot state $Q_{r,r} = diag([0.1, 0.0001])$.

The operator's target is shown on a monitor, with the robot one and the robot tip's current position. The human is asked to reach the target position in physical interaction with the robot. Such a target position is not visible on the monitor. The human only knows the weight $\alpha = \{0.2, 0.5, 0.9\}$ (defined offline). The target poses are $z_{ref,h} = 0.6m$ and $z_{ref,r} = 0.3m$.

The robot applies a virtual force $u_r$, while the human applies a real, measured force to the system. The balance of the two forces is turned into a velocity command through an impedance controller. The human and the robot move the system toward their goals (which are different for NCGT and LQR, the agreed one for CGT), knowing each other's goals.

Figure 5 plots the state and efforts for three controllers with $\alpha = 0.5$. As from 5(a-b-c), the equilibrium solution is reached in all three cases. The human measured and nominal control efforts are in 5(d-e-f). The nominal control effort is computed with the gain matrix $K_{h,i} = R_h^{-1} B_h^T P_{h,i}$ with $i = \{cgt, lqr, ncgt\}$. As expected from simulations, CGT shows that once the equilibrium (i.e., agreed reference) is reached, neither the human nor the robot puts any more effort into the system. On the contrary, the other two controllers still inject effort into the system to keep the equilibrium. Moreover, this continuous pulling of the robot towards its reference introduces oscillations, as the human does not exactly behave as modeled. On the contrary, CGT appears more smooth. Humans and robots share the same target and cooperatively steer the system towards it without "pulling" it in different directions. The human costs of performing real-world experiments are computed for a time window of 3.5s. The results are in figure 6. This measurement reflects what is observed in the simulated scenarios: the human incurs high costs for low values of $\alpha$, and vice versa, low costs for high values. The opposite happens for the LQR and NCGT because it is more convenient to work with high values of $\alpha$ for humans. In such a case, a lower effort produces better results. High values of $\alpha$ lead to a shared target close to the human's, and the robot drives the system toward it.

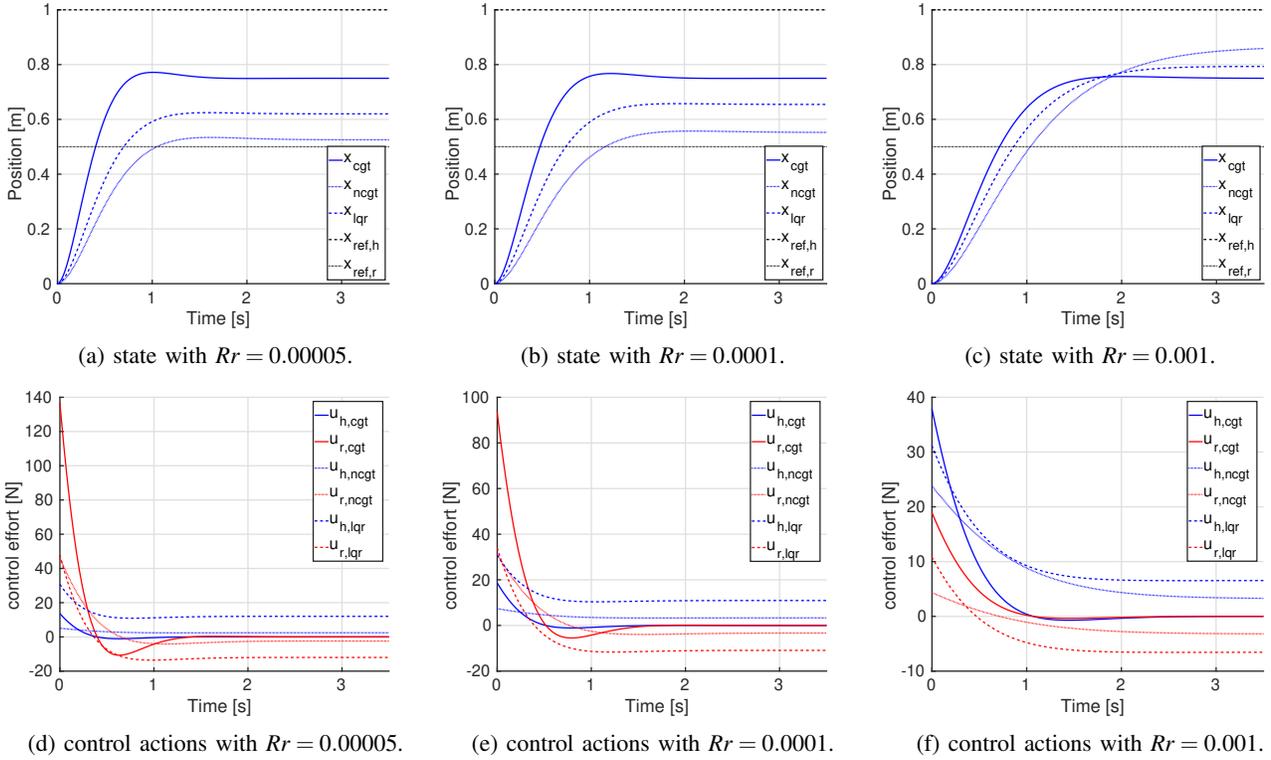

Fig. 3: Position and control efforts according to different robot weight on control action $R_r$.

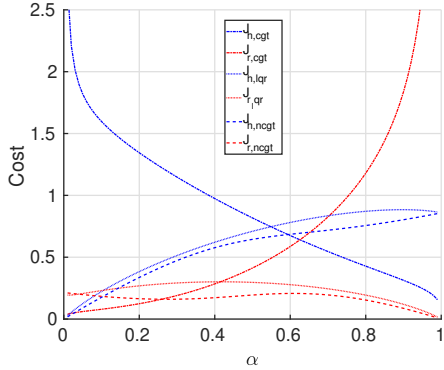

Fig. 4: Human (blue lines) and robot (red lines) costs for the three controllers for various $\alpha$ in the simulated cases.

Finally, the error between measured and nominal human control efforts is computed for the three controllers for three values of $\alpha$. The errors are the sum over the time windows of 3.5s and normalized by the maximum nominal control input:

$$\mathscr{E}_u = \frac{1}{max(u_{h,n})} \int_{t_i}^{t_f} \|u_h - u_{h,n}\| dt \quad (18)$$

Figure 7 shows that the CG better fits with the pHRI, making it more applicable to real-world applications.

## IV. DISCUSSION

Consider that a high value of $\alpha$ corresponds to the human-leading, while a low value corresponds to the robot-leading. In the real world, a human likely wants the help of the robot to pursue his target but would probably not accept putting too much effort into the task to help the robot when the $\alpha$ is low. In this sense, a scenario suggests using CGT when the human leads and NCGT or LQR when the robot leads. Therefore, the robot helps the human in the first case and drives the system toward its reference when it wants to lead.

As simulations show, a small robot's control action weight increases the robot's reactiveness, helping humans. For high values of $\alpha$, the robot is more willing to help the human reach its target, while for low values, the robot is more willing to reach its target and can exert higher forces compared with the case in which $R_r = R_h$. This behavior could make cooperation between humans and robots more comfortable.

Real-world experiments confirm that for the CGT case, the human cost is higher for low values of $\alpha$ and vice versa for the other two. Comparing the three real-world experiments, CGT fits the model better, allowing modeling the pHRI as a cooperative game, which might be helpful for simulations and tuning of the controllers. Moreover, having a model that reasonably fits with data possibly allows for online prediction of human behavior. This prediction capability can improve assistive controllers' design and provide the robot with some information about the future behavior of its opponents.

Limitations to real-world applications are mainly due to human modeling. First, the experiments are performed by a trained human, aware of CGT methods, while operators do

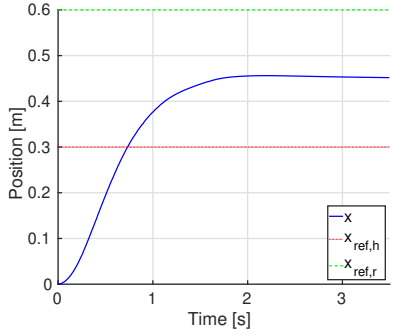
(a) measured position for CG with $\alpha = 0.5$.

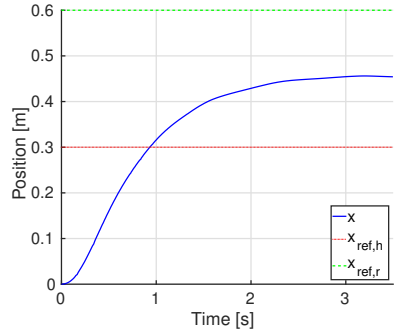
(b) measured position for LQR with $\alpha = 0.5$.

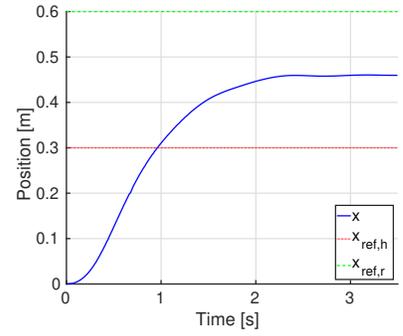
(c) measured position for NCG with $\alpha = 0.5$.

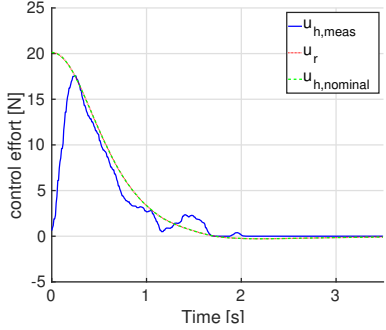
(d) CG measured human and robot control efforts and nominal human control effort.

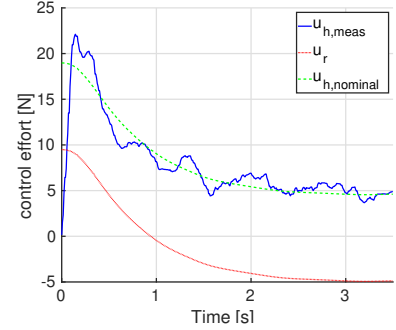
(e) LQR measured human and robot control efforts and nominal human control effort.

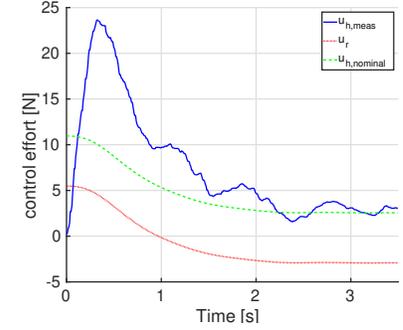
(f) NCG measured human and robot control efforts and nominal human control effort.

Fig. 5: position and control effort comparison for the three controllers with the weight parameter $\alpha = 0.5$

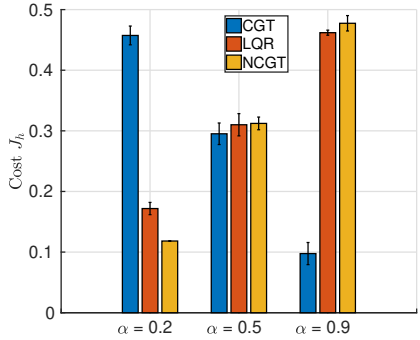
Fig. 6: Computed Human cost varying $\alpha$.

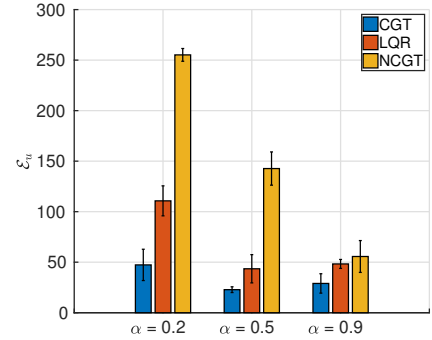
Fig. 7: human's control errors for the three controllers varying $\alpha$.

not precisely follow the CGT formulation. Second, human costs vary from person to person and possibly along the work shift. Third, the task could make it vary. For this purpose, the presented method should be enriched with a robustness term. Furthermore, it could be possible to investigate the performance of online IOC techniques to identify the cost function parameters or the human control gains in run-time [30]. Finally, as stated above, this work takes the human target for granted. This is typically not true in real-world applications, and predictive models for the human target should be investigated and integrated.

## V. CONCLUSIONS

This work presented a controller for the pHRI based on Cooperative Game Theory. A shared reference is computed, and a solution to the LQ-CGT reference tracking problem is obtained. Starting from two different references, the controller converges to an agreed one, sharing the required effort to drive the system to the target. The proposed method is compared in simulations and real-world experiments with the two controllers, LQR and NCGT. A set of simulations investigates the performances of the proposed controller, varying the control parameters. Real-world experiments partially confirm what was observed in the simulated cases and show that CGT better fits with the pHRI task. Finally, a discussion of the results obtained is proposed, addressing the possibility and limitations of the proposed methodology. Future works will involve online human-robot role arbitration, making variable the parameter $\alpha$ in run-time, possibly switching between CGT and LQR/NCGT. Finally, the control scheme will include online human target position prediction, as it is crucial for real-world applications.

## APPENDIX - CGT MATRICES COMPUTATION

In this section, the computations of $Q_{gt}$, $R_{gt}$ and $z_{ref}$ are detailed. Rewriting and rearranging (5) and (6) leads to

$$J_h = \int_0^\infty \big( z^T (Q_{h,h} + Q_{h,r}) z + z_{ref,h}^T Q_{h,h} z_{ref,h} + z_{ref,r}^T Q_{h,r} z_{ref,r}$$
$$- 2 z_{ref,h}^T Q_{h,h} z - 2 z_{ref,r}^T Q_{h,r} z + u_h^T R_h u_h \big) \, dt$$
$$J_r = \int_0^\infty \big( z^T (Q_{r,h} + Q_{r,r}) z + z_{ref,h}^T Q_{r,h} z_{ref,h} + z_{ref,r}^T Q_{r,r} z_{ref,r}$$
$$- 2 z_{ref,h}^T Q_{r,h} z - 2 z_{ref,r}^T Q_{r,r} z + u_r^T R_r u_r \big) \, dt \quad (19)$$

Combining these equations into (7), it results in

$$J_{gt} = \int_0^\infty \big( z^T Q_{gt} z + z_{ref,h}^T Q_h z_{ref,h} + z_{ref,r}^T Q_r z_{ref,r} +$$
$$- 2 z_{ref,h}^T Q_h z - 2 z_{ref,r}^T Q_r z + u^T R_{gt} u \big) \, dt \quad (20)$$

with

$$Q_{gt} = \alpha (Q_{h,h} + Q_{h,r}) + (1-\alpha)(Q_{r,h} + Q_{r,r}),$$
$$Q_h = \alpha Q_{h,h} + (1-\alpha) Q_{r,h}, Q_r = \alpha Q_{h,r} + (1-\alpha) Q_{r,r}$$
$$\text{and} \quad R_{gt} = \begin{bmatrix} \alpha R_h & 0 \\ 0 & (1-\alpha) R_r \end{bmatrix}.$$

Using (7), $z_{ref}$ can be computed by:

$$J_{gt} = \int_0^\infty \big((z - z_{ref})^T Q_{gt} (z - z_{ref}) + u^T R_{gt} u\big) \, dt$$
$$= \int_0^\infty \big(z^T Q_{gt} z + z_{ref}^T Q_{gt} z_{ref} - 2 z_{ref}^T Q_{gt} z + u^T R_{gt} u\big) \, dt$$
$$(21)$$

Being constant $z_{ref,h}^T Q_h z_{ref,h}$, $z_{ref,r}^T Q_r z_{ref,r}$ and $z_{ref}^T Q_{gt} z_{ref}$, the solution of problem (9) is not affected, and the only components in (20) and (21) to be compared are

$$-2 z_{ref}^T Q_{gt} z = -2 (z_{ref,h}^T Q_h + z_{ref,h}^T Q_r) z \quad (22)$$

Simplifying, $z_{ref}$ results in

$$z_{ref} = Q^{-1}(z_{ref,h} Q_h + z_{ref,h} Q_r) \quad (23)$$

## ACKNOWLEDGMENT

The research has been partially funded by the EU project Drapebot. Grant agreement no 101006732. The research has been partially funded by the EU project FLUENTLY. Grant agreement no 101058680.